\documentclass[runningheads]{llncs}
\usepackage{graphicx}

\usepackage{tikz}
\usepackage{comment}
\usepackage{amsmath,amssymb} 
\usepackage{color}
\usepackage[pagebackref,breaklinks,colorlinks]{hyperref}
\usepackage[capitalize]{cleveref}

\usepackage[accsupp]{axessibility}  

\usepackage{graphicx}
\usepackage{amsmath}
\usepackage{amssymb}
\usepackage{booktabs}

\usepackage{tabularx}
\usepackage{bm}
\usepackage{wrapfig}
\usepackage{subfigure}
\usepackage{multirow}
\usepackage{ulem}
\usepackage{caption}

\usepackage{CJK}
\usepackage{algorithm}
\usepackage{algorithmicx}
\usepackage{algpseudocode}

\begin{document}
\pagestyle{headings}
\mainmatter
\def\ECCVSubNumber{4636}  

\title{Learning Quality-aware Dynamic Memory for Video Object Segmentation} 

\titlerunning{Learning Quality-aware Dynamic Memory for Video Object Segmentation}
\author{Yong Liu\inst{1*}\and
Ran Yu\inst{1} \and
Fei Yin\inst{1} \and
Xinyuan Zhao\inst{2} \and
Wei Zhao\inst{2} \and
Weihao Xia\inst{3} \and
Yujiu Yang\inst{1}$^{\dagger}$
}
\authorrunning{Y. Liu et al.}
%
\institute{Tsinghua Shenzhen International Graduate School, Tsinghua University \and
Huawei Technologies \and
University College London \\
\email{\{liu-yong20,yu-r19\}@mails.tsinghua.edu.cn}, \email{yang.yujiu@sz.tsinghua.edu.cn}}
\maketitle

\let\thefootnote\relax\footnotetext{\scriptsize{$^*$This work was done during an internship at Huawei Technologies}}
\let\thefootnote\relax\footnotetext{\scriptsize{$^{\dagger}$Corresponding author}}

\begin{abstract}
 
Recently, several spatial-temporal memory-based methods have verified that storing intermediate frames and their masks as memory are helpful to segment target objects in videos. However, they mainly focus on better matching between the current frame and the memory frames without explicitly paying attention to the quality of the memory. Therefore, frames with poor segmentation masks are prone to be memorized, which leads to a segmentation mask error accumulation problem and further affect the segmentation performance. In addition, the linear increase of memory frames with the growth of frame number also limits the ability of the models to handle long videos. To this end, we propose a \textbf{Q}uality-aware \textbf{D}ynamic \textbf{M}emory \textbf{N}etwork (QDMN) to evaluate the segmentation quality of each frame, allowing the memory bank to selectively store accurately segmented frames to prevent the error accumulation problem. Then, we combine the segmentation quality with temporal consistency to dynamically update the memory bank to improve the practicability of the models. Without any bells and whistles, our QDMN achieves new state-of-the-art performance on both DAVIS and YouTube-VOS benchmarks. Moreover, extensive experiments demonstrate that the proposed Quality Assessment Module (QAM) can be applied to memory-based methods as generic plugins and significantly improves performance.
Our source code is available at \textcolor{magenta}{\url{https://github.com/workforai/QDMN}}.

\end{abstract}

\section{Introduction}

Given a video and the first frame's annotations of single or multiple objects, semi-supervised video object segmentation (Semi-VOS or One-shot VOS) aims at segmenting these objects in subsequent frames. Semi-VOS is one of the most challenging tasks in computer vision with many potential applications, including interactive video editing, augmented reality, and autonomous driving. 

\begin{figure}[t]
  \centering
  \includegraphics[width=0.98\linewidth]{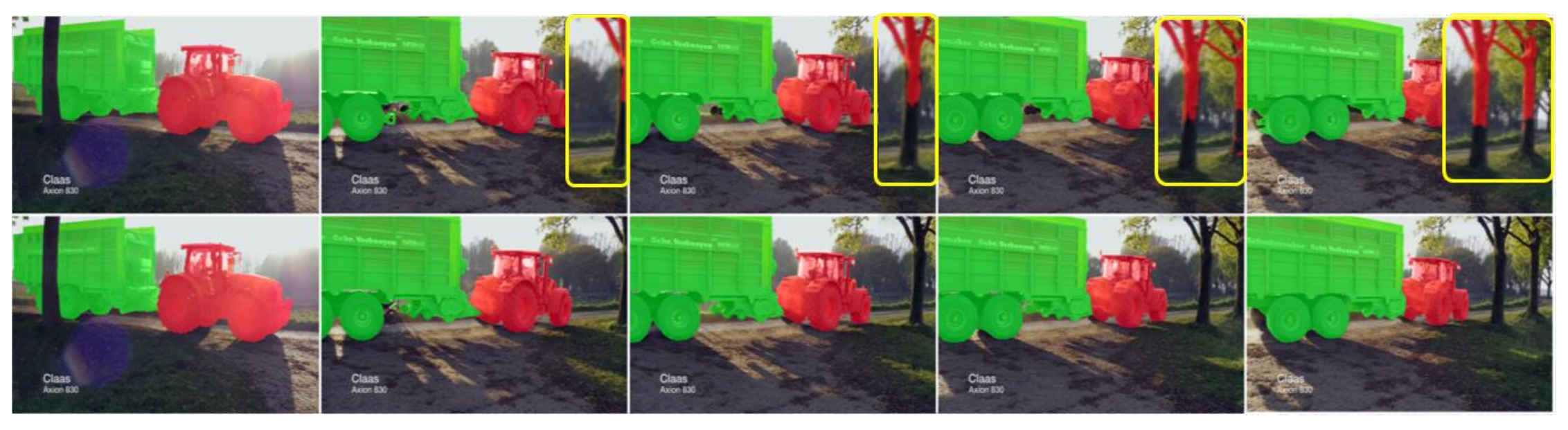}
  \caption{Visual comparison of memory frames of different qualities. 
  The first row shows the memory frames of MiVOS~\cite{mivos}. The second row shows the memory frames of our method. The yellow box area illustrates the error accumulation.}
  \label{bank}
\end{figure}
Unlike other segmentation tasks that aim to look for the relationship between features and specific categories, the critical problem of Semi-VOS lies in how to make full use of the spatial-temporal information to recognize the target objects. 
Consequently, the methods that perform matching with historical reference frames have received tremendous attention in recent years.
Some works~\cite{feelvos,cfbi,cfbi+} utilize the first frame and the previous adjacent frame as references.
Due to limited reference information, these approaches tend to fail miserably under challenging scenarios, \textit{e.g.}, the target objects disappear for a while or are drastically deformed.
To excavate more information, the Space-Time Memory Network (STM)~\cite{stm} utilizes a memory network to memorize intermediate frames and their segmentation masks as references, which has been proved effective and has served as the current mainstream framework.
Many approaches~\cite{kmn,gcm,mivos,lcm,rmnet,hmm,swiftnet,stcn} further develop the feature extraction and memory readout process of STM and have achieved excellent performance.

However, these methods mainly focus on optimizing the matching process while ignoring the impact of the matching target, \textit{i.e.,} memory bank, on the segmentation results.
Specifically, previous methods select memory frames in a straightforward way, \textit{i.e.}, storing at fixed frame intervals. 
This approach has two weaknesses: 
(1) Frames with poor segmentation results may be memorized and provide an erroneous reference for subsequent frames, which leads to an error accumulation problem.
As shown in the first row of \cref{bank}, if there are inaccurately segmented masks in the memory bank, the segmentation quality of subsequent frames will be greatly degraded. 
Such an observation inspires us to pay more attention to the design of the memory bank.
Since the matching-based approaches rely on a memory bank to identify the target objects, the memory bank's quality (especially the correctness) is very important.
(2) In existing methods, the size of the memory bank would infinitely expand with the growth of frame number, which makes the models incapable of handling long videos and greatly limits their practicality.

Therefore, the way of designing the memory bank is a significant issue for spatial-temporal memory-based methods.
Generally speaking, we believe that the design of the memory bank should meet the following principles: (1) \textit{\textbf{Accuracy:} In a one-shot scenario, the memory bank should be composed of the annotated frame and frames that are segmented as accurately as possible to obtain correct supervision information.} 
(2) \textit{\textbf{Temporal consistency:} Considering the continuity of motion, the state of objects in adjacent frames tends to be similar. In other words, the masks of adjacent frames are of great reference to the current frame.} Based on these two principles, we can selectively store frames with more reference information as memory and dynamically update the memory bank to handle videos of arbitrary length. 

To this end, we propose a Quality-aware Dynamic Memory Network (QDMN), which
introduces a simple but effective structure called Quality Assessment Module (QAM) in this paper to evaluate each frame's segmentation result and decide whether a frame can be added to the memory bank as a reference.
Being aware of the segmentation quality limits the impact of noise and provides the accuracy credentials for dynamically updating the memory bank.
Besides, since the objects in adjacent frames share a similar status to the current target, we introduce a temporal regularization to penalize the outdated memory.
Extensive experiments demonstrate that the dynamic updating strategy of the memory bank designed according to the principles of accuracy and temporal consistency is reasonable and effective.
By designing a high-quality memory bank and introducing temporal consistency, our method achieves new state-of-the-art performance on both DAVIS~\cite{davis17} and Youtube-VOS~\cite{youtube} benchmark without any bells and whistles. 
Furthermore, we also verify that memory-based methods can gain significant improvement by simply applying our QAM as a generic plugin for video object segmentation tasks.


Our contributions can be summarized as follows.
Firstly, we pinpoint the design of the memory bank as the Achilles heel of the Semi-VOS task and propose the strategy for designing a high-quality memory bank.
Secondly, we present QDMN for Semi-VOS, which can selectively memorize high-quality frames and take advantage of the temporal consistency.
Thirdly, QDMN can effectively control the number of memory frames to avoid memory explosion.
Experiments show that our method surpasses the existing methods on both DAVIS and YouTube-VOS datasets. Furthermore, QAM can be used as a generic plugin to improve memory-based methods.

\section{Related Work}

    \subsubsection{Propagation-based Methods.} Propagation-based methods~\cite{objectflow,segflow,favos,dvsnet,osvos,maskrnn,onavos} treat semi-supervised video object segmentation as a mask propagation task.
    MaskTrack~\cite{masktrack} concatenates the previous adjacent frame's segmentation mask with the current image as input and online fine-tunes the network. 
    AGSS-VOS~\cite{agss-vos} proposes an attention-guided decoder to combine the instance-specific branch and instance-agnostic branch.
    Based on mask confidence and mask concentration, SAT~\cite{sat} selectively propagates the entire image or local region to the next frame.
    The propagation-based method takes advantage of the strong prior provided by the previous adjacent frame. It can better deal with the appearance change of the target object, but it has fatal shortcomings in the problem of occlusion and error accumulation.
    
    \subsubsection{Detection-based Methods.} Detection-based methods divide the Semi-VOS task into three subtasks: detection, tracking and segmentation.
    DyeNet~\cite{dyenet} utilizes RPN~\cite{fasterrcnn} to generate proposals and applies the re-identification module to perform matching. 
    PReMVOS~\cite{premvos} uses Mask RCNN~\cite{maskrcnn} to obtain coarse masks and performs optical flow, re-identification to achieve good performance. 
    Huang \textit{et al}~\cite{topdown1} and Sun \textit{et al}~\cite{topdown2} integrate segmentation into tracking with a dynamic template bank. 
    Detection-based methods rely heavily on the detectors, which dramatically limits the performance of such methods.
    
    \subsubsection{Matching-based Methods.} 
    Matching-based methods perform matching between reference frames and the current frame to identify target objects, which has raised great attention for excellent performance and robustness.
    PML~\cite{pml} proposes a pixel-level embedding network with the nearest neighbor classifier.
    FEELVOS \cite{feelvos} and CFBI~\cite{cfbi} perform global and local matching with the first frame and the previous adjacent frame, respectively.
    AOT~\cite{aot} associates multiple target objects into the same embedding space by employing an identification mechanism.
    STM~\cite{stm} leverages the memory network to memorize intermediate frames as references, which has been proved effective and has served as the current mainstream framework. 
    Based on STM, KMN~\cite{kmn} and RMNet~\cite{rmnet} propose to perform local-to-local matching instead of non-local.
    SwiftNet~\cite{swiftnet} and AFB-URR~\cite{afb-urr} reduce memory duplication redundancy by calculating the similarity between query and memory.
    LCM~\cite{lcm} emphasizes the importance of the first frame and the previous adjacent frame.
    STCN~\cite{stcn} improves the feature extraction and performs reasonable matching by decoupling the image and masks.
    Following the memory-based idea, there are still many variants of STM, such as JOINT~\cite{joint}, EGMN~\cite{egmn}, MiVOS~\cite{mivos}, DMN-AOA~\cite{alignment}, HMMN~\cite{hmm}, and so on.
    
    Although these methods have achieved great performance, they mainly focus on better matching the current frame with the memory frames.
    In other words, previous works dedicate to optimizing the matching process while neglecting the importance of matching with the correct object.
    Besides, they do not take into account that the size of the memory bank grows linearly with the length of the video, which greatly impacts the application of the models in real scenarios due to the hardware memory limitation.
\section{Method}

\begin{figure*}[t]
    \centering
    \includegraphics[width=\textwidth]{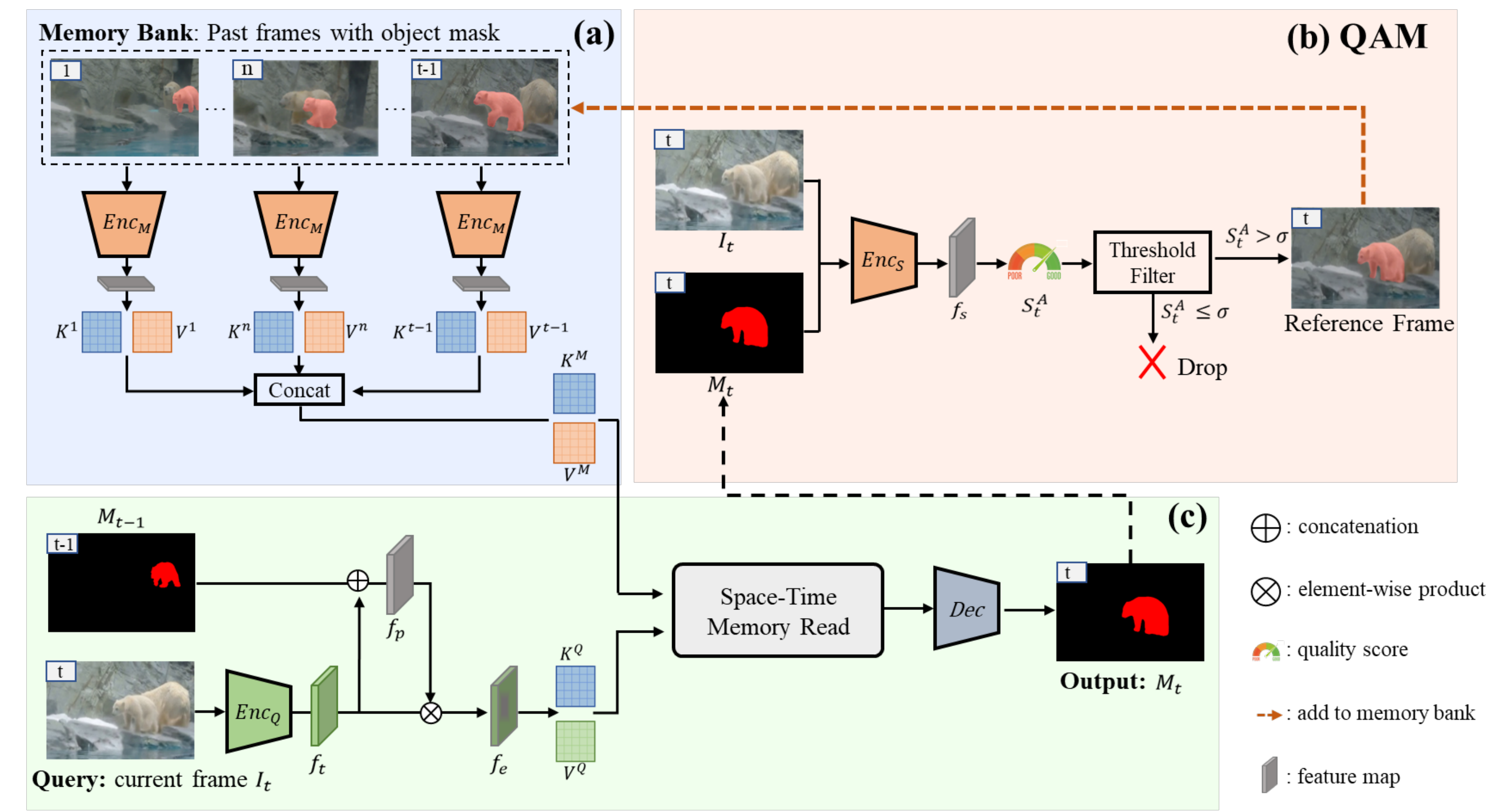}
    \caption{Overview of QDMN. (a) is the feature extraction of the reference frames in the memory bank. (b) QAM is the module used to evaluate whether the current frame can be added to the memory bank. (c)
    is the pipeline for predicting the segmentation result of the current frame ${I}_t$ .
    }
    \label{framework}
\end{figure*}


\subsection{Overview}\label{overview}
    The overall architecture of our QDMN is shown in \cref{framework}. 
    Similar to STM~\cite{stm}, during video processing, the current frame ($t$-th frame) is considered as the query, and the past reference frames with segmentation masks are considered as the memory.
    The query and memory are encoded into pairs of key and value maps through visual encoders and corresponding convolution layers.
    To highlight the temporal consistency of video, the query feature $f_t$ is first enhanced with the prior mask to obtain the enhanced feature $f_e$. Then the enhanced feature is encoded into pairs of key ${K}^Q$ and value ${V}^Q$ through corresponding convolution layers.
    The Space-Time Memory Read block performs pixel-level matching between ${K}^Q$ and the memory key ${K}^M$.
    The relative matching similarity is used to address the memory value ${V}^M$, and the corresponding values are combined to the decoder for segmentation.
    Finally, the Quality Assessment Module (QAM) evaluates the quality of the segmentation result and decides whether the query frame can become a memory frame.
        \begin{figure*}[t]
      \centering
      \includegraphics[width=\textwidth]{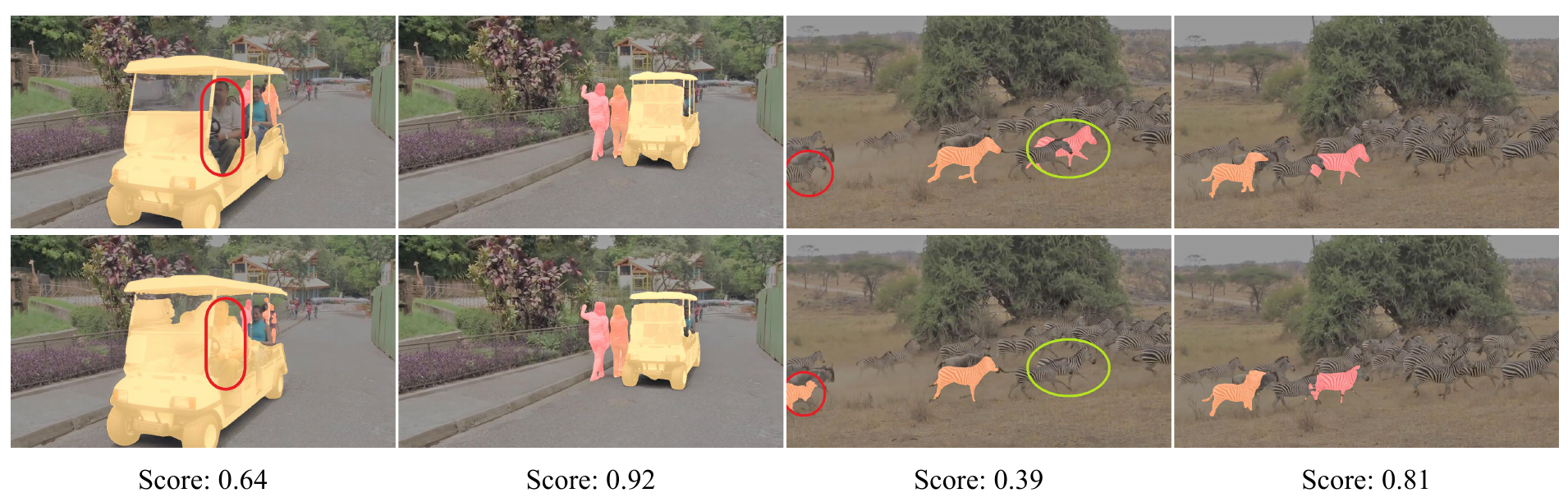}
      \caption{Illustrations of segmentation masks with different quality scores. 
      The three rows represent the ground truth, segmentation results, and the quality scores predicted by QAM, respectively.}
      
      \label{figure score}
    \end{figure*}

\subsection{Quality Assessment Module}\label{QAM}
    Designing the memory bank is a significant issue for memory network-based methods. 
    For existing strategy, frames with erroneous masks may be memorized, which leads to an error accumulation problem.
    To alleviate this problem and ensure the accuracy of the memory bank, inspired by ~\cite{msrcnn,iou-net}, we propose the Quality Assessment Module (QAM) to evaluate the segmentation quality and decide whether a frame can be added to the memory bank as a reference.
    
    QAM is a simple structure but effective module composed of a score encoder, four convolution layers, and two MLP layers. 
    It takes the query image $\mathbf{I}_t$ and its segmentation mask $\mathbf{M}_t$ as input and outputs the predicted quality scores.
    Since the feature extraction process of the score encoder $Enc_{s}$ is the same as that of the memory encoder $Enc_{M}$ (both takes images with segmentation masks as input), we directly use the memory encoder as the score encoder, which helps to save calculations and parameters. 
    Specifically, the structure of the score encoder $Enc_{s}$ and the memory encoder $Enc_{M}$ is the same, and the parameters are shared. 
    The QAM first takes the query image $\mathbf{I}_t\in\mathbb{R}^{3\times H\times W}$ and its segmentation mask $\mathbf{M}_t\in\mathbb{R}^{1\times H\times W}$ into the score encoder to obtain the score feature map $f_s\in\mathbb{R}^{C\times H/16\times W/16}$,
    where $H\times W$ are resolutions of the input image. 
    Then, $f_s$ is input to the  convolution layers and fully connected layers to learn the segmentation quality score $\mathbf{S}^A_t$ for the current frame. The process of segmentation quality assessment can be expressed as:
    \vspace{-7pt}
    \begin{equation}
        f_s = {Enc}_s(\mathbf{I}_t \oplus \mathbf{M}_t); \quad
        \mathbf{S}_t^A = \mathit{Fc}(\mathit{Conv}(f_s)),
    \end{equation}
    where $\oplus$ denotes the concatenation operation. 
    $t$ is the index of the current frame.
    $\mathit{Conv}$  and $\mathit{Fc}$ denote convolution and fully connected layers with sigmoid non-linear function, respectively. 
    
    During training, the target value of the quality score is defined as mask IoU between the segmentation mask and ground truth. 
    The specific calculation process is as follows:
    \begin{equation}
        \textit{loss} = \frac{1}{N} \sum^N_{i=1}(S^A_i-maskIoU(M_i,{GT}_i))^2,
    \end{equation}
    where $S^A_i$ represents the quality score of the segmentation result for $i$-th object, $M_i$ indicates the segmentation result, ${GT}_i$ is the ground truth. $N$ indicates the total number of objects.

    Since QAM evaluates the segmentation quality for each object individually, we take the average of all object scores in one frame as the quality score of this frame. In addition, considering that the segmentation difficulty varies for different video scenes, we normalize the quality scores of all frames in a video to better measure the relative quality of the segmentation results, which helps to memorize more helpful information under challenging scenarios. Specifically, the final quality score of each frame is its initial predicted score divided by the score of the first frame. Formally, the process can be expressed as:

    \begin{equation}
        \mathbf{\bar{S}}_t^A = \frac{\frac{1}{N}\sum^N_{i=1}\mathbf{S}_{t_i}^A}{\mathbf{\bar{S}}_1^A},
    \end{equation}
    where N represents the total number of objects in the t-th frame, $\bar{S}_t^A$ indicates the quality score of the segmentation result in frame $t$, $\bar{S}_1^A$ represents the quality score of the first frame.

    \Cref{figure score} shows some visualization results of the quality assessment, the first two columns are the same video, and the last two columns represent another video. 
    We can observe that the driver is considered part of the car in the first column, which is a bad case.
    The pink zebra in the third column is not recognized, and the orange zebra is matched with similar background objects. 

    For the hard case,
    our QAM identifies these suboptimal results well, which shows that the segmentation accuracy of a frame is consistent with its quality score. 
    Extensive experiments also verify this.
    With QAM, the memory bank can selectively memorize frames whose quality scores are higher than the memory threshold $\sigma$, that is, frames with accurate segmentation masks.
    In this way, even if a frame is poorly segmented owing to
    fast object motion or other factors, 
    it will not affect the subsequent frames or cause error accumulation. 
    
    	\begin{wrapfigure}{r}{5cm}
		\begin{minipage}[p]{1\linewidth}
            \centering
			\includegraphics[width=\linewidth,height=1.5\linewidth]{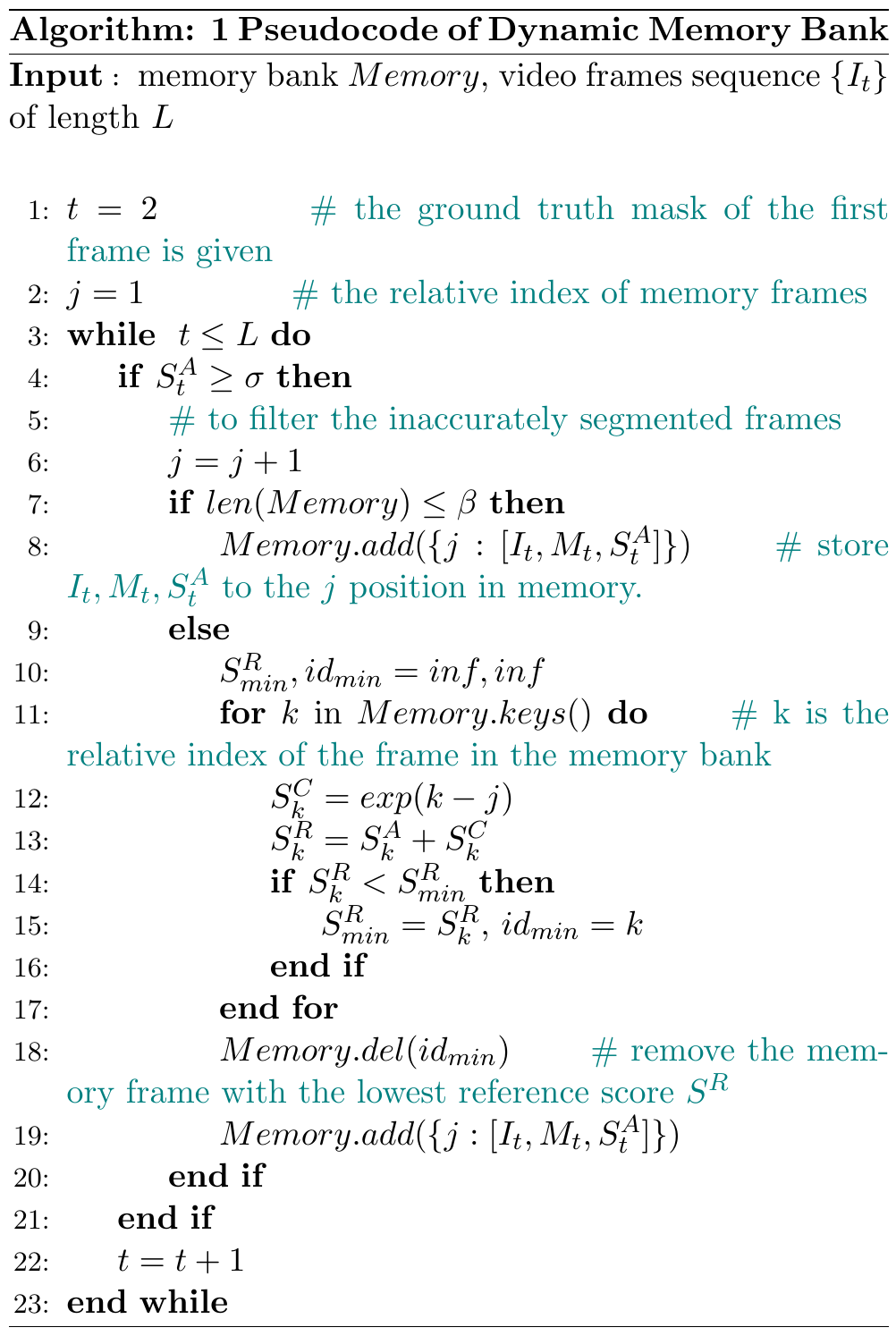}
			\label{memory cap}
		\end{minipage}
	\end{wrapfigure} 
    
    \subsection{Dynamically Updated Memory Bank}
     
    The infinite increase of the memory frames with the growth of frame number greatly limits the practicability of the model in the real-world scenario.
    Thus, it is necessary to limit the size of the memory bank and update it dynamically to adapt to new scenarios.
    
    Due to the temporal consistency of video, the appearance of the target objects in adjacent frames is similar. 
    The masks of adjacent frames are more instructive for the segmentation of current frame.
    Combining the above analysis and considering accuracy, we suggest dynamically updating the memory bank in accordance with these two principles (Algorithm.1).
    Specifically, when the memory bank reaches a certain storage limit, we will dynamically update the memory bank to handle different video scenes.
    For quantifying the temporal consistency and measuring the distance between each memory frame and the current frame, we compute the temporal consistency score $\mathbf{S}^C$ as:
    \begin{equation}
    \label{consistency}
        \mathbf{S}_k^C = e^{-|t-k|},
    \end{equation}
    where $k$ is the index of each memory frame, $t$ is the index of the current frame.
    
    Based on the accuracy score $\mathbf{S}^A$ and the temporal consistency score $\mathbf{S}^C$, the reference score of each memory frame in the memory bank can be calculated by $\mathbf{S}^R_k = \mathbf{\bar{S}}^A_k + \mathbf{S}^C_k$.
    By removing the memory frames with the lowest reference score, the memory bank is dynamically updated to handle different video scenarios and prevent the memory explosion problem.

    \subsection{Prior Enhancement Strategy} 
    
    In addition to considering temporal consistency when designing a memory bank, we further utilize the prior provided by the previous adjacent frame to enhance temporal information.
    We adopt a similar module structure to SCM~\cite{davis1st} to introduce the prior information from the previous adjacent frame.
    Instead of introducing spatial constraint in the decoder like SCM, we utilize the prior information in the query encoding process to better learn the target object's appearance feature and avoid over-reliance on the prior information.
    
    Specifically, in the query encoding process, the segmentation mask of the previous adjacent frame $\mathbf{M}_{t-1}\in\mathbb{R}^{1\times H\times W}$ is downsampled and concatenated with the query's embedding $f_t\in\mathbb{R}^{C\times H/16\times W/16}$.
    Then the resultant feature goes through convolution and non-linear function to fuse information between channels, through which a prior feature map $f_p\in\mathbb{R}^{1\times H/16\times W/16}$ is produced.
    Finally, we perform an element-wise product between $f_{p}$ and $f_t$ to get the enhanced feature $f_e\in\mathbb{R}^{C\times H/16\times W/16}$. Formally, the process can be expressed as the following equation:
    \begin{equation}
        f_{e} = \mathit{Conv}(f_t \oplus \mathbf{M}_{t-1}) \otimes f_t.
    \end{equation}

    Furthermore, we find that it is better to provide weak prior (mentioned above) than strong prior (masks of the previous frame have a great influence on the feature of the current frame).
    We found two primary reasons through experiments: the first one is that the prior information may be noisy, and providing a strong prior may lead to error accumulation; 
    the second one is that providing strong prior makes the model overly dependent on it, which weakens its ability to extract features and identify objects. 
    \cref{table strategy} shows the disadvantages of providing strong prior under challenging scenarios. 
    In \cref{ablation study}, we will describe the specific approach of providing strong prior.

\subsection{Memory Read and Decoder}
    
    In the Space-Time Memory Read block~\cite{stm}, soft weights are first computed by measuring the similarities between query key $K^Q$ and memory key $K^M$. Then the memory value $V^M$ is retrieved by a weighted summation with the soft weights and concatenated with query value $V^Q$ to get the output $y$. This operation can be summarized as:
    \begin{equation}
        y_i = V^Q_i \oplus \frac{1}{Z}\sum_{\forall j}{\mathcal{D}(K^Q_i,K^M_j)V^M_j},
    \end{equation}
    where $i$ and $j$ are the index of the query and the memory location, $Z = \sum_{\forall j}\mathcal{D}({K}^Q_i,{K}^M_j)$ is the normalizing factor. $\mathcal{D}$ denotes the similarity measure (in our experiment is dot product).
    
    Our decoder stays close to that of \cite{decoder,stm}. The decoder takes the output $y$ of the Space-Time Memory Read block as input and predicts the object masks. It consists of an ASPP layer~\cite{aspp}, a residual block, and two upsample blocks that upscale the feature map to the initial image size.

\section{Implementation Details}
    Following the training strategy in MiVOS~\cite{mivos}, we first pretrain our model on static image datasets~\cite{static1,static2,static3,static4,static5} and then perform main training on YouTube-VOS and DAVIS datasets. Besides, we also experiment with the synthetic dataset BL30K proposed in MiVOS, which is not used unless otherwise specified. 
    During pretraining, each image is expanded into a pseudo video of three frames by random affine, horizontal flip, color and brightness augmentation. 
    We randomly pick three frames in chronological order (with a ground-truth mask for the first frame) from a video to form a training sample in the main training.
    The range of random sampling varies with the training process.
    In the intermediate period of training, the sampling range is set larger to improve the robustness of the model, while at the end of the training, it is set smaller to narrow the gap between training and inference.
    Our models are trained end-to-end with two 32GB Tesla V100 GPUs with the Adam optimizer in PyTorch. The batch size is set to 28 during pretraining and 16 during main training.
    We adopt ResNet-50~\cite{resnet} as backbone for all encoders. Bootstrapped cross-entropy loss~\cite{mivos} is used for segmentation, and MSE loss is used for quality score evaluation. 
    The initial learning rate is 2e-5.
    During inference, we choose the memory threshold $\sigma$ of 0.8 by default. 
    Ablation studies are conducted on a single 1080Ti GPU and DAVIS 2017 validation set in default.
    
\section{Experiments}
	\begin{table*}[t]
		\centering
		\caption{Comparison with other methods on DAVIS dataset. `*' indicates using synthetic training dataset~\cite{mivos}.}
		\label{tab:results}
		\setlength\tabcolsep{4pt}
		\begin{tabular}{lccccccccc}
			\toprule[1.5pt]
			\multirow{2}{*}{Method} & \multicolumn{3}{c}{DAVIS2016} & \multicolumn{3}{c}{DAVIS2017 val} & \multicolumn{3}{c}{DAVIS2017 test-dev} \\
			\cmidrule(lr){2-4} \cmidrule(lr){5-7} \cmidrule(lr){8-10}
			& $\mathcal{J}$ & $\mathcal{F}$ & $\mathcal{J}\&\mathcal{F}$ & $\mathcal{J}$ & $\mathcal{F}$ & $\mathcal{J}\&\mathcal{F}$& $\mathcal{J}$ &$\mathcal{F}$ & $\mathcal{J}\&\mathcal{F}$ \\
			\midrule
			RANet~\cite{ranet}   & 86.6 & 87.6 & 87.1 & 63.2  & 68.2 & 65.7 & 53.4    & 56.2    & 55.3 \\
			FEELVOS~\cite{feelvos}  & 81.1 & 82.2 & 81.7 & 69.1   & 74.0   & 71.5 & 55.2    & 60.5    & 57.8\\
			RGMP~\cite{rgmp}      & 81.5   & 82.0 & 81.8 &64.8 &68.6 &66.7 &51.3 &54.4 &52.8\\
			DMVOS~\cite{dmvos}   & 88.0  & 87.5  & 87.8 &- &- &- &- &- &-\\
			STM~\cite{stm}       & 88.7  & 89.9  & 89.3 & 79.2   & 84.3   & 81.8 & 69.3    & 75.2    & 72.2\\
			KMN~\cite{kmn}       & 89.5  & 91.5 & 90.5 & 80.0    & 85.6    & 82.8 & 74.1    & 80.3    & 77.2\\
			CFBI~\cite{cfbi}     & 88.3  & 90.5  & 89.4 & 79.1    & 84.6    & 81.9 & 71.1    & 78.5    & 74.8\\
			GIEL~\cite{giel}     &- &- &- & 80.2    & 85.3    & 82.7 & 72.0    & 78.3    & 75.2\\
			SwiftNet~\cite{swiftnet} & 90.5  & 90.3  & 90.4 & 78.3    & 83.9    & 81.1 &- &- &-\\
			RMNet~\cite{rmnet}   & 88.9  & 88.7 & 88.8 & 81.0    & 86.0    & 83.5 & 71.9    & 78.1    & 75.0\\
			SSTVOS~\cite{sstvos}     &- &- &-  & 79.9  & 85.1  & 82.5   &- &- &- \\ 
			LCM~\cite{lcm}       & 89.9  & 91.4  & 90.7 & 80.5    & 86.5    & 83.5 & 74.4    & 81.8    & 78.1\\
			MiVOS~\cite{mivos}   & {87.8}   & {90.0}  & {88.9} & {80.5}    & {85.8}    & {83.1} &{72.6}    & {79.3}    & {76.0}\\
			MiVOS*~\cite{mivos}  & {89.7}  & {92.4} & {91.0} & {81.7} & {87.4} & {84.5} & {74.9} & {82.2} & {78.6}\\
			JOINT~\cite{joint}   &- &- &-  & 80.8 & 86.2  & 83.5   &- &- &-  \\
			RPCMVOS~\cite{aaai} &87.1 &94.0 &90.6 &81.3 &86.0 & 83.7 &75.8 &82.6 &79.2\\
			DMN-AOA~\cite{alignment}     &- &- &- & 81.0   & 87.0 & 84.0  &74.8 &81.7 &78.3  \\
			HMMN~\cite{hmm}       &89.6 &92.0 &90.8 &81.9 &87.5 &84.7  & 74.7  & 82.5  & 78.6  \\
			STCN~\cite{stcn}    & 90.8 & 92.5 & 91.6 & 82.2 & 88.6 & 85.4 & 72.7 & 79.6 & 76.1\\
			AOT-L~\cite{aot}  & 89.7  & 92.3 & 91.0 & 80.3   & 85.7 & 83.0 & 75.3   & 82.3 & 78.8\\
			\midrule
			QDMN (Ours)  & \textbf{90.2}  & \textbf{91.7} & \textbf{91.0} & \textbf{81.8}  & \textbf{87.3} & \textbf{84.6} & \textbf{74.2}  & \textbf{81.2}  & \textbf{77.7}\\
			QDMN* (Ours)           & \textbf{90.7}  & \textbf{93.2} & \textbf{92.0} & \textbf{82.5}  & \textbf{88.6} & \textbf{85.6} & \textbf{78.1}  & \textbf{85.4}  & \textbf{81.9}\\
			\bottomrule[1.5pt]
		 \vspace{-25pt}
		\end{tabular}
		
	\end{table*}

\subsection{Comparisons with State-of-the-Art Methods}

    \noindent\textbf{DAVIS 2016}~\cite{davis16} is a single object benchmark for video object segmentation. As shown in \Cref{tab:results}, QDMN trained without synthetic dataset still outperforms most previous methods (\textbf{91.0} $\mathcal{J\&F}$).
    With synthetic training data, QDMN surpasses all existing methods and achieves the performance of \textbf{92.0} $\mathcal{J\&F}$.

    \noindent\textbf{DAVIS 2017}~\cite{davis17} is a multiple objects extension of DAVIS 2016. In the \Cref{tab:results}, QDMN achieves an average score of \textbf{84.6} and \textbf{85.6} for training without synthetic data and with synthetic data, respectively. What's more, we also test our model on the challenging DAVIS 2017 testing split set. It achieves the best performance (\textbf{81.9}) compared to all previous methods.
    
    \begin{table}[t]
    \centering
    \caption{Evaluation on YouTube-VOS 2018 val set. 
    Seen and Unseen denote whether the categories exist in the training set.
    $\mathcal{G}$ is averaged overall score.}
    \label{youtube}
    \setlength\tabcolsep{11pt}
    \begin{tabular}{lccccc}
        \toprule[1.5pt]
        \multirow{2}{*}{Methods} &
        \multicolumn{2}{c}{Seen} & \multicolumn{2}{c}{Unseen} &
        \multirow{2}{*}{$\mathcal{G}$} \\
        \noalign{\smallskip} \cline{2-5} \noalign{\smallskip}
              & $\mathcal{J}$ & $\mathcal{F}$
              & $\mathcal{J}$ & $\mathcal{F}$ \\
        \midrule
    STM~\cite{stm}          & 79.7    & 84.2  & 72.8    & 80.9  & 79.4          \\
    AFB-URR~\cite{afb-urr}          & 78.8    & 83.1  & 74.1    & 82.6  & 79.6          \\
    GCM~\cite{gcm}        & 72.6    & 75.6  & 68.9    & 75.7  & 73.2          \\
    KMN~\cite{kmn}          & 81.4    & 85.6  & 75.3   & 83.3  & 81.4          \\
    G-FRTM~\cite{g-frtm}    & 68.6    & 71.3  & 58.4   & 64.5  & 65.7          \\
    SwiftNet~\cite{swiftnet} & 77.8   & 81.8  & 72.3   & 79.5  & 77.8          \\
    GIEL~\cite{giel}         & 80.7   & 85.0  & 75.0   & 81.9  & 80.6          \\
    SSTVOS~\cite{sstvos}     & 80.9   & -     & 76.6     & -     & 81.8          \\
    RMNet~\cite{rmnet}       & 82.1   & 85.7  & 75.7   & 82.4  & 81.5          \\
    LCM~\cite{lcm}           & 82.2 & 86.7 & 75.7  & 83.4    & 82.0          \\
    MiVOS~\cite{mivos}       & {80.0}    & {84.6} & {74.8}   & {82.4}  & {80.4}          \\
    MiVOS*~\cite{mivos}      & {81.1}    & {85.6} & {77.7}  & {86.2} & {82.6}          \\
    JOINT~\cite{joint}                 & 81.5 & 85.9 & 78.7 & 86.5 & 83.1\\
    HMMN~\cite{hmm}                 & 82.1 & 87.0 & 76.8 & 84.6 & 82.6\\
    DMN-AOA~\cite{alignment}                 & 82.5 & 86.9 & 76.2 & 84.2 & 82.5\\
    STCN~\cite{stcn}                 & 81.9 & 86.5 & 77.9 & 85.7 & 83.0\\
    AOT-L~\cite{aot}  & 82.5   & 87.5 & 77.9 & 86.7 & 83.7 \\
    \midrule
    QDMN (Ours)      & \textbf{82.0}  & \textbf{86.8}  & \textbf{77.5}   & \textbf{85.5} & \textbf{83.0} \\
    QDMN* (Ours)     & \textbf{82.7}  & \textbf{87.5}  & \textbf{78.4}   & \textbf{86.4}               & \textbf{83.8} \\
    \bottomrule[1.5pt]
    \end{tabular}
    \end{table}
    
    \noindent \textbf{YouTube-VOS}~\cite{youtube} is a large-scale benchmark for video object segmentation.
    As shown in \Cref{youtube}, without synthetic training data, our QDMN also achieves state-of-the-art performance (\textbf{83.0}). If we use synthetic data for training, the overall score of QDMN will be boosted to \textbf{83.8}.

\begin{figure*}[t]
    \centering
    \includegraphics[width=\textwidth]{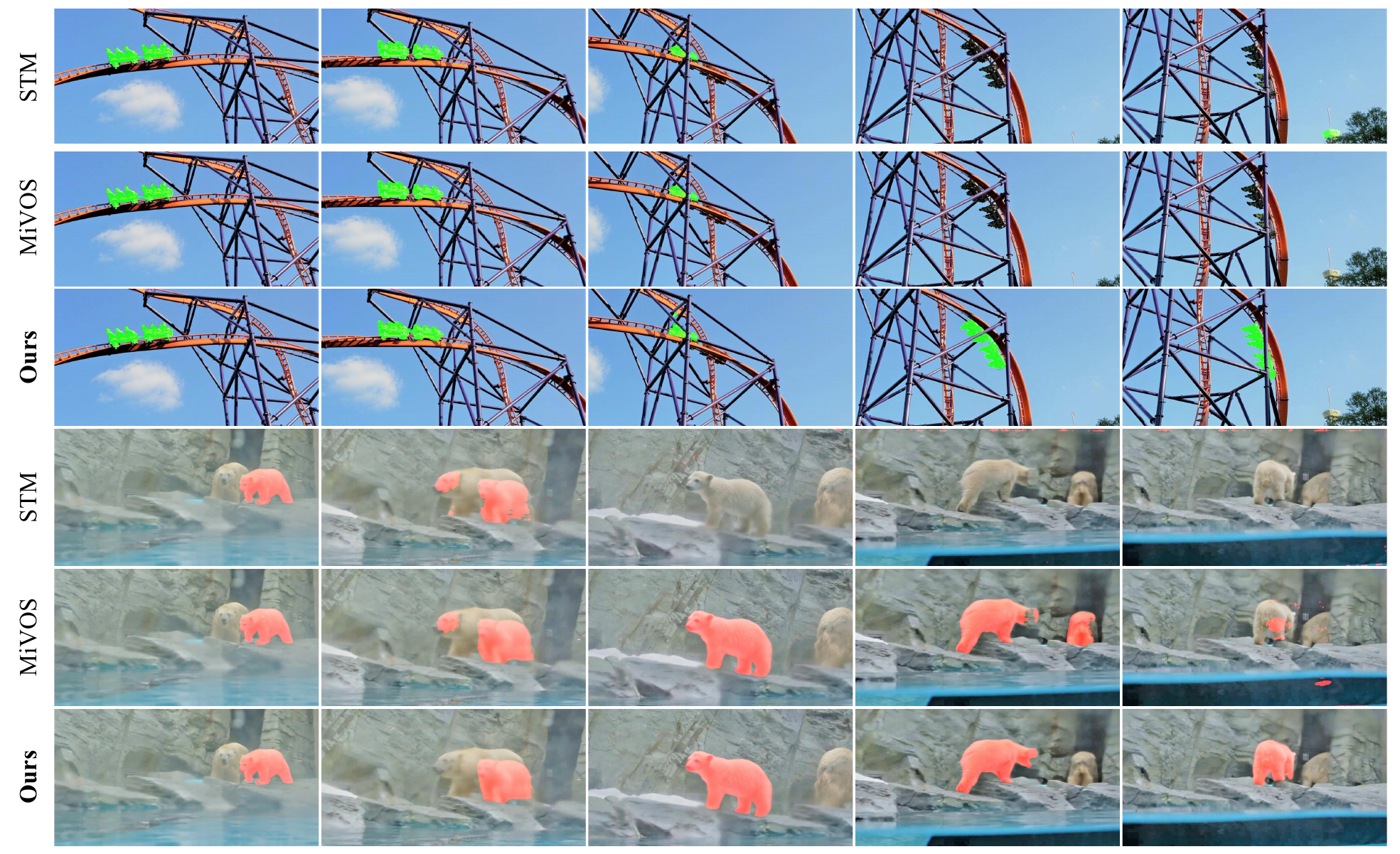}
    \caption{Visual comparison of QDMN with baseline methods.Each row demonstrates five frames sampled from a video sequence.}
    \label{compare result}
\end{figure*}

    \noindent \textbf{{Qualitative results.}} The qualitative comparison between baseline and our QDMN are shown in \cref{compare result}. We show the performance on two challenging scenarios, \textit{i.e.}, occlusion scenes and similar objects. 
    Both STM~\cite{stm} and MiVOS~\cite{mivos} have lost targets in the occlusion scene. STM lost targets in the scene with similar objects, while MiVOS identified other objects incorrectly. In contrast, our method can achieve satisfactory performance in challenging scenarios.

    \subsection{Generic Plugins}
\begin{figure}[h]
    \centering
    \begin{minipage}[b]{0.54\linewidth}
        \includegraphics[width=1.0\linewidth]{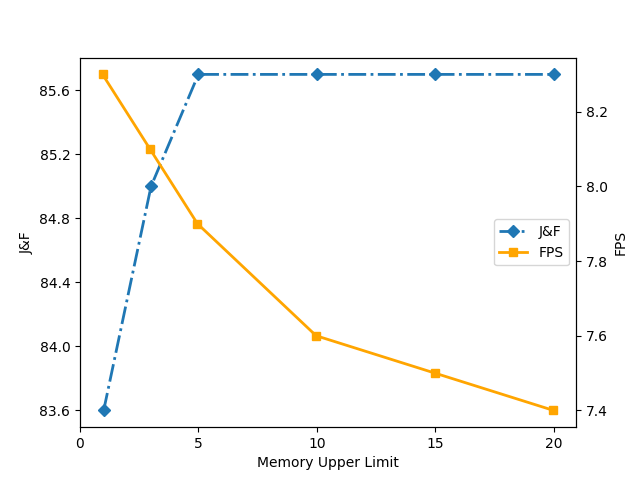}
        \caption{The performance for different memory upper limit.}
        \label{memory cap}
    \end{minipage}
    \hfill
    \begin{minipage}[b]{0.44\linewidth}
    \centering
    \small
        \begin{tabular}{lcccc}
    \toprule[1.5pt]
                    Methods    & w / QAM & $\mathcal{J}$             & $\mathcal{F}$      & $\mathcal{J}\&\mathcal{F}$       \\ \midrule
    \multirow{2}{*}{STM~\cite{stm}}         &            & 78.8          & 84.2    & 81.5    \\      
                                 &\checkmark  & 81.0 & 86.2 & \textbf{83.6}$^{\uparrow}$\\ \midrule
    \multirow{2}{*}{KMN~\cite{kmn}}         &          & 79.7          & 85.5     & 82.6      \\   
                                 &\checkmark  & 81.9 & 87.4 & \textbf{84.7}$^{\uparrow}$ \\ \midrule
    \multirow{2}{*}{STCN~\cite{stcn}}        &           & 81.5          & 87.7     & 84.6    \\   
                                 &\checkmark     & 82.5 & 88.7    & \textbf{85.6}$^{\uparrow}$     \\ 
    \bottomrule[1.5pt]
    \end{tabular}
    \captionof{table}{Applying QAM as general plugin. w / QAM indicates that whether the QAM is deployed on this method.}
    \label{plug}
    \end{minipage}
\end{figure}

    To further prove the effectiveness of our proposed QAM, we apply it as a general plugin to other methods. 
    The results on the DAVIS2017 validation set are shown in \cref{plug} (the baseline performance is our re-implementation results). 
    It can be seen that with QAM, the performance of these methods has been significantly boosted. 
    Besides, QAM is easy to be deployed on other methods, and we hope that the QAM would shed light on the studies of related fields that need to memorize reference information.

\subsection{Ablation Study}\label{ablation study}

    \subsubsection{The effectiveness of QAM.}
    To demonstrate the effectiveness of the QAM, we conduct specific analyses from three dimensions.

\begin{figure}[h]
    \tiny
    \centering
    \begin{minipage}[b]{0.48\linewidth}
        \includegraphics[width=1.0\linewidth]{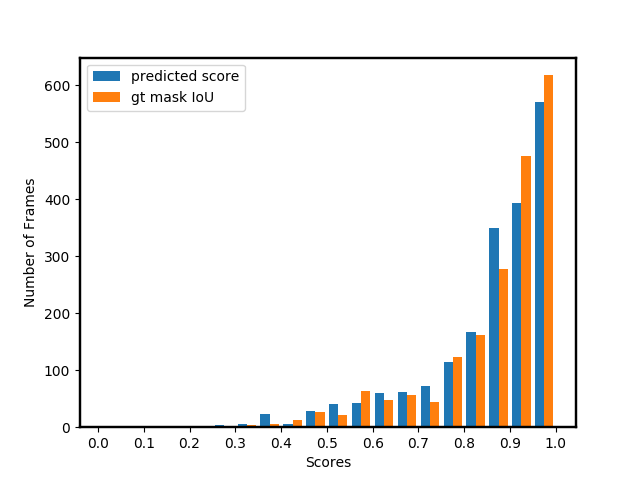}
        \caption{Distribution of the prediction score and the ground truth mask IoU.}
		\label{fig:distribution}
    \end{minipage}
    \hfill
    \begin{minipage}[b]{0.48\linewidth}
        \includegraphics[width=1.0\linewidth]{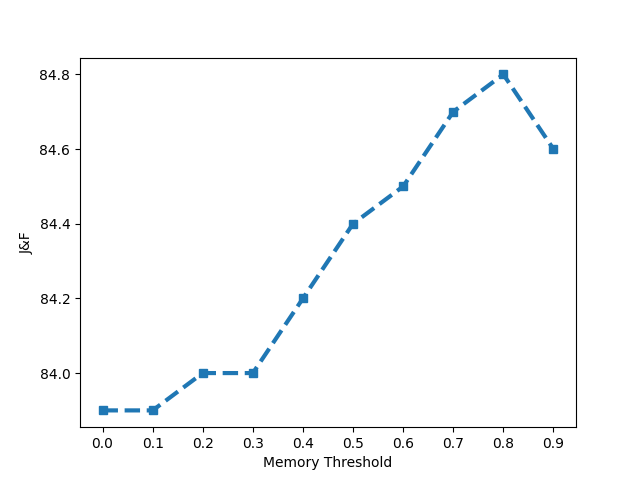}
        \caption{The quantitative results of different memory threshold $\sigma$.}
        \label{fig:memory_threshold}
    \end{minipage}
\end{figure}

    \textbf{(1) Accuracy of the predicted scores.}
    We perform a histogram visualization of the distribution of the ground truth mask IoU and prediction scores at 0.05 intervals(\cref{fig:distribution}).
    When multiple objects are in a frame, the average is taken. 
    We can see that the quality score and ground truth mask IoU are positively correlated, which verifies the accuracy of the scores predicted by QAM.
    
    \textbf{(2) Memory Threshold.} We test different memory thresholds $\sigma$ on DAVIS 2017 test-dev set, and the results are shown in \cref{fig:memory_threshold}. 
    We can see that it will hurt the segmentation effect if the threshold is set too high or too low.
    The reason is that if the threshold $\sigma$ is too high, only a few intermediate frames will be memorized, leading to losing a lot of helpful information; if the $\sigma$ is too low, the model may memorize some incorrect noise information.
    Besides, the performance is worst when the memory threshold is 0 (at this time, QAM does not filter poor segmentation masks), which proves the motivation of the QAM is correct. 
    
    
    \textbf{(3) Applying QAM only at inference stage.}
    To further prove that filtering out inaccurately segmented frames has a beneficial effect on segmentation, 
    we construct experiments that adding QAM only at the inference stage. 
    Specifically, for QAM, we load its parameters trained in QDMN.
    For other parts, we load the weights of the initial model (trained without QAM).
    As shown in~\cref{tab:onlyinference}, the performance of all vanilla models has been improved after adding QAM, which shows the importance of filtering poorly segmented frames.\\
    
\begin{minipage}{\textwidth}
 \begin{minipage}[h]{0.45\textwidth}
  \centering
     \makeatletter\def\@captype{table}\makeatother
     \caption{The effect of adding QAM only in \textit{the inference stage}}
     \label{tab:onlyinference}
     \renewcommand\arraystretch{1.25}
     \renewcommand\tabcolsep{4pt}
     \begin{tabular}{lccc}
    \toprule[1.5pt]
    Methods      &$\mathcal{J}\&\mathcal{F}_{(\sigma = 0)}$     &$\mathcal{J}\&\mathcal{F}_{(\sigma = 0.8)}$    \\ \hline
    STM                            & 81.5  & \textbf{82.5$\uparrow$}    \\      
    KMN                         & 82.6      & \textbf{83.4$\uparrow$}\\   
    MiVOS                  & 82.7    & \textbf{83.5$\uparrow$}\\   
    \bottomrule[1.5pt]
    \end{tabular}
  \end{minipage}
  \begin{minipage}[h]{0.05\textwidth}
  \ 
  \end{minipage}
  \begin{minipage}[h]{0.43\textwidth}
  \centering
        \makeatletter\def\@captype{table}\makeatother\caption{Ablation study of proposed components.}
        \footnotesize
        \label{component analysis}
        \renewcommand\tabcolsep{3pt}
    \begin{tabular}{ccccc}
    \toprule[1.5pt]
    QAM     &PEM  & $\mathcal{J}$ & $\mathcal{F}$ & $\mathcal{J}\&\mathcal{F}$ \\ 
    \midrule
    &     & 80.3  &  85.5     & 82.9   \\ 
    \checkmark  &   & 81.7  & 87.1 & \textbf{84.3}$^{\uparrow}$   \\
    & \checkmark & 81.1 & 86.1 & \textbf{83.6}$^{\uparrow}$    \\
    \checkmark  & \checkmark & 81.8 & 87.3 & \textbf{84.6}$^{\uparrow}$   \\
    \bottomrule[1.5pt]
    \end{tabular}
   \end{minipage}
\end{minipage}

    
    \subsubsection{Component Analysis.}
    We analyze the effectiveness of our modules in~\cref{component analysis}. 
    PE represents the prior enhancement strategy introduced to highlight temporal consistency.
    As shown in the table, both the QAM and PE bring remarkable performance improvement.
    
    \subsubsection{Dynamic Memory Updating Strategy.} Due to the lack of a widely used large-scale long video dataset in this field, we choose to demonstrate the effectiveness of our proposed memory bank dynamic updating strategy by compressing the upper limit of the memory.
    As shown in \Cref{memory cap}, The segmentation effect remains unaffected even at low memory upper limit, and the speed is improved
    as a result of our memory bank design strategy.
    The similar phenomenon is observed on the YouTube-VOS set, which illustrates the effectiveness of our dynamic updating strategy.
    
    Besides, we also perform analysis on long videos (without annotations).
    We find that previous memory network 
    methods store up to about 70 frames and the memory explosion occurs, 
    which greatly limits the practicability. 
    But QDMN can handle videos of arbitrary length by setting upper memory limit and dynamically updating the memory.
    What's more, the FPS of previous methods will drop from 14 to about 2 before memory explodes, while the FPS of QDMN will stay around 7 after the initial drop (assuming the upper memory limit is 25).
    
    \begin{wraptable}{r}{6cm}
    \centering
    \caption{Ablation study of different enhancement strategy. 
    ``Weak" means providing weak prior (PE). 
    ``Strong" means providing strong location prior.}
    \label{table strategy}
    \begin{tabular}{lcccccc}
    \toprule[1.5pt]
            
            \multirow{2}{*}{Strategy} &\multicolumn{3}{c}{DAVIS} &\multicolumn{3}{c}{YouTube-VOS}\\
        \cmidrule(lr){2-4} \cmidrule(lr){5-7}
              & $\mathcal{J}$  & $\mathcal{F}$  & $\mathcal{J}\&\mathcal{F}$  & $\mathcal{J}$  & $\mathcal{F}$  & $\mathcal{G}$\\
              \midrule
    
    Weak              & 81.8    & 87.3 & 84.6 & 79.8 & 86.2 & \textbf{83.0}    \\        \noalign{\smallskip} 
    Strong              & 82.4  & 87.9  &\textbf{85.2}   & 77.5 & 83.8 & 80.7 \\
    \bottomrule[1.5pt]
    \end{tabular}
    \end{wraptable}

    \subsubsection{Enhancement Strategy.} 
    For PE, we directly concatenate the prior mask with the deepest layer feature of the current frame to provide a weak prior. In contrast, we also try to provide a strong prior. Specifically,  we extract the feature of the prior mask and fuse it with the middle layer features of the current frame. After convolution and downsampling, the fused features are added to the deepest layer features of the current frame. Compared with the current enhancement strategy, this approach can significantly enhance the influence of the prior mask. However, although this approach works well in common scenarios, the performance drops significantly under challenging situations, as shown in \cref{table strategy}.
    The reason for this phenomenon is that the strong prior makes the model overly dependent on it, which weakens the model's ability to recognize objects.

    \subsubsection{Speed Analysis.} We also experiment with the impact of the proposed modules on the inference speed. With our modules, the FPS of baseline has changed from \textbf{8.6} to \textbf{7.8} on DAVIS2017 val set. 
    The increased running time brought by QAM and PE is nearly negligible (no more than 10$\%$), mainly because we directly use the feature extracted by the memory encoder for quality assessment.

\section{Conclusion}
    In this paper, we propose that the design of the memory bank should follow the principles of accuracy and temporal consistency.
    To support this, we introduce a Quality-aware Dynamic Memory Network (QDMN) for semi-supervised video object segmentation, which selectively memorizes accurately segmented intermediate frames as references and emphasizes video temporal consistency.
    Without bells and whistles, our QDMN achieves new state-of-the-art performance on the popular benchmark YouTube-VOS and DAVIS with almost no additional inference time. Furthermore, the QAM also has a remarkable improvement for other approaches as a general plugin.

\subsection*{Acknowledgments.}
This research was supported in part by the National Natural Science Foundation of China under Grant No. U1903213, the Shenzhen Key Laboratory of Marine IntelliSense and Computation (NO. ZDSYS20200811142605016.)

%
%
\bibliographystyle{splncs04}
\bibliography{egbib}

\newpage
\appendix
\section*{Appendix}

\section{Analysis on Long Videos}
\begin{figure}[h]
      \centering
      \includegraphics[width=\textwidth]{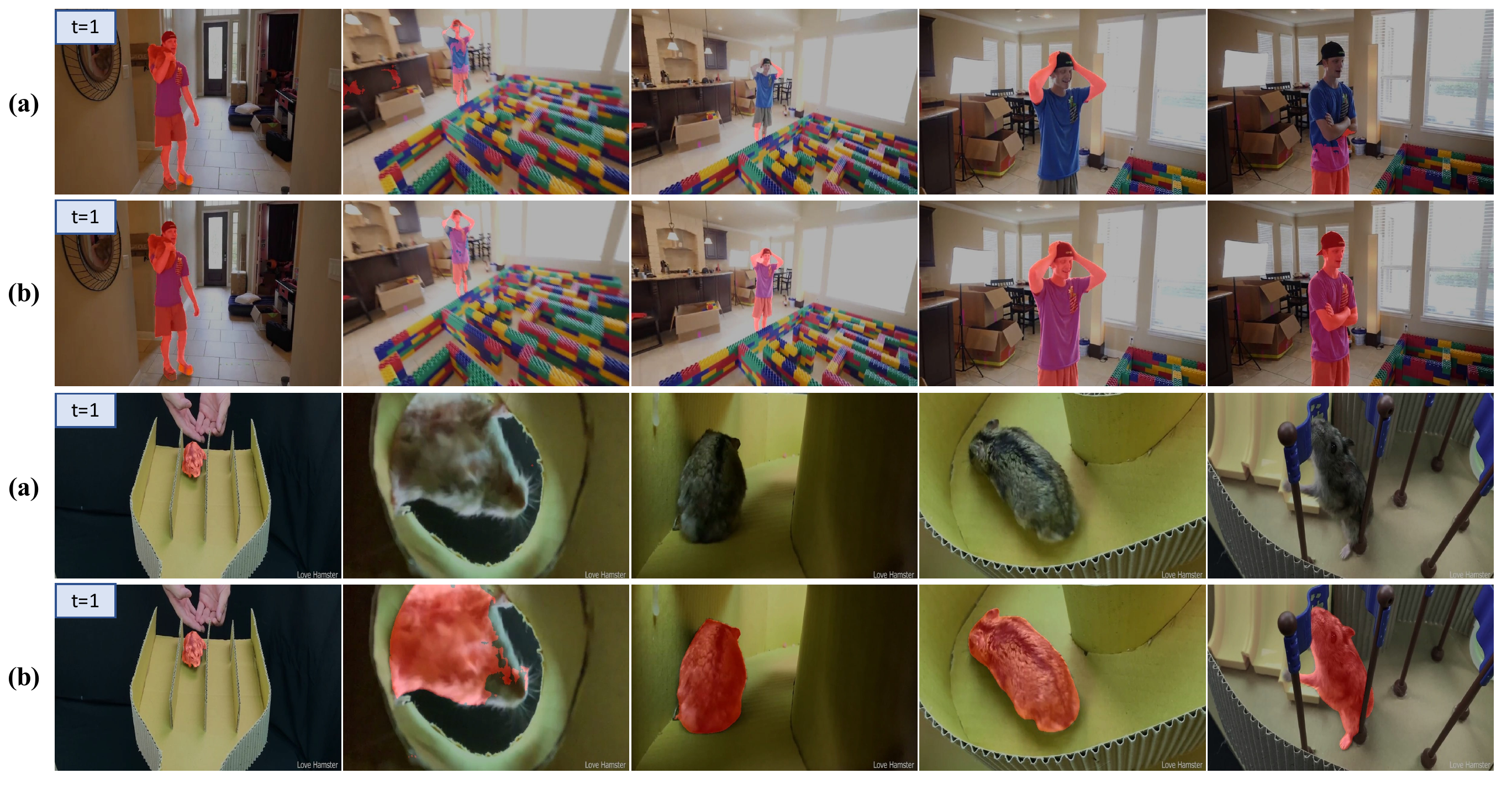}
      \caption{Qualitative results of our proposed dynamic memory updating strategy. (a) is the results of retaining the most recent memory frames and (b) is applying our updating strategy. The memory frame storage limit is 25 frames.}
      \label{long_video}
    \end{figure}

To further prove the rationality and effectiveness of our dynamic memory updating strategy, we show the qualitative results on long videos (more than 2000 frames) in~\cref{long_video}.
In general, the most common practice for updating memory in practical applications is to retain the most recent memory frames.
However, this approach is difficult to deal with object appearance changes or scene changes.
As shown in~\cref{long_video}, only retaining the most recent memory frames may cause the memory bank losing perception of the target object but our dynamic updating strategy allows for superior segmentation effect.

\section{Failure Cases}
\begin{figure}[t]
      \centering
      \includegraphics[width=\textwidth]{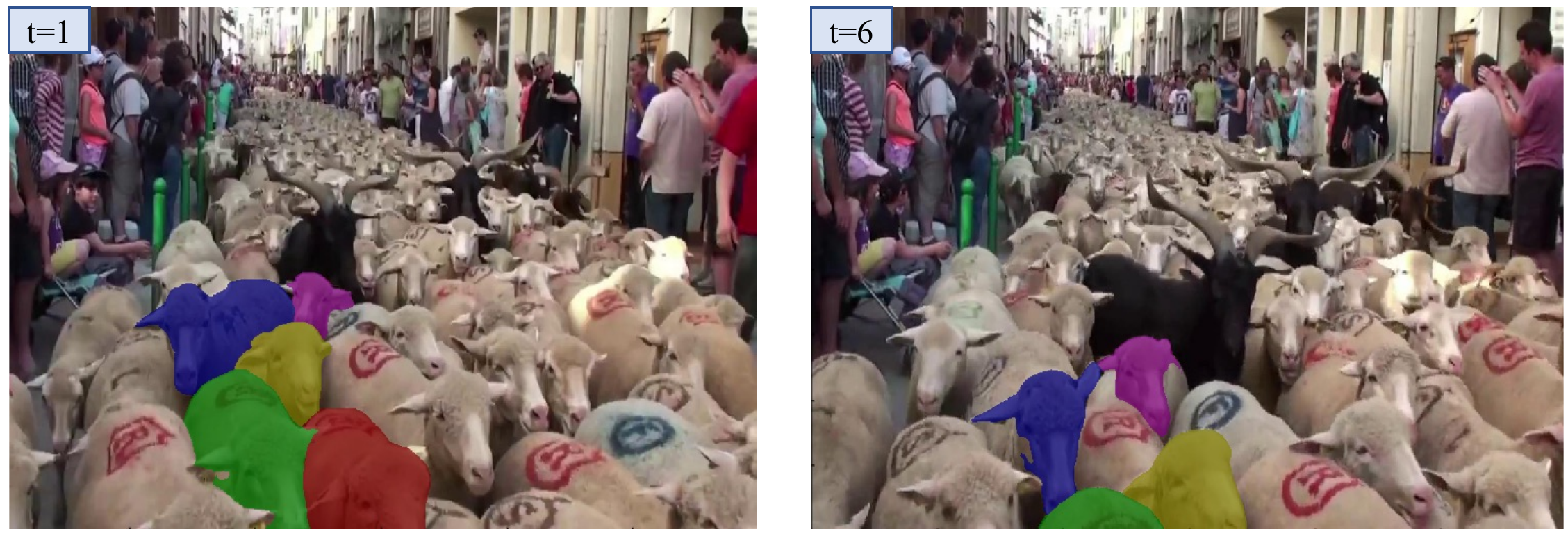}
      \caption{The bad case of QAM.}
      \label{bad_case}
    \end{figure}

The failure cases of QAM are the extremely difficult scenarios, \textit{e.g.}, lots of similar objects overlapping each other, in which almost every frame is poorly segmented.
Take the \cref{bad_case} as example, in this case, the target sheep are mixed with other background sheep.
Along with the movement of the flock, it is difficult for the algorithm to correctly identify the target sheep as well as the outline of the sheep.
Although we use relative quality scores in this scenario, QAM is less helpful.

\section{Memory Interval}
    In order to avoid excessive memory redundancy in the same scene, 
    we still choose to trigger the storage of every N frames, but the frame must meet our proposed principles. 
    Otherwise, it will be deferred to the next frame to trigger. 
    As \Cref{memory interval} shows, it is not the smaller the memory interval, the better the segmentation effect. 
    The reason is that although the smaller memory interval means more reference frames, excessive redundancy in the same scene affects the matching process.
    \begin{table}[h]
    \renewcommand\tabcolsep{5pt}
    \footnotesize
    \centering
    \caption{Experiment results of different memory interval on DAVIS2017 val set.}
    \label{memory interval}
    \begin{tabular}{cccc}
    \toprule[1.5pt]
    Memory Interval & $\mathcal{J}$ &$\mathcal{F}$ & $\mathcal{J}\&\mathcal{F}$ \\ \midrule
    3               & 82.3 & 88.1 & 85.2 \\
    5               & \textbf{82.7} & \textbf{88.6} & \textbf{85.7} \\
    7               & 81.6 & 87.8 & 84.7 \\ \bottomrule[1.5pt]
    \end{tabular}
    \end{table}


\end{document}


\pagestyle{headings}
\mainmatter
\def\ECCVSubNumber{4636}  

\title{Learning Quality-aware Dynamic Memory for Video Object Segmentation} 

\maketitle

\section{Experiments}
\paragraph{\textbf{Analysis on long videos.}}
To further prove the rationality and effectiveness of our dynamic memory updating strategy, we show the qualitative results on long videos in~\cref{dynamic-memory}.
In general, the most common practice for updating memory in practical applications is to retain the most recent memory frames.
However, this approach is difficult to deal with object appearance changes or scene changes.
As shown in~\cref{dynamic-memory}, only retaining the most recent memory frames may cause the memory bank losing perception of the target object.
And our dynamic updating strategy allows for superior segmentation effect while controlling the memory frame storage limit.
    \begin{figure}[h]
      \centering
      \includegraphics[width=\textwidth]{figure/dynamic_bank.pdf}
      \vspace{-10pt}
      \caption{Qualitative results of our proposed dynamic memory updating strategy. (a) is the results of retaining the most recent memory frames and (b) is applying our updating strategy. The memory frame storage limit is 25 frames.}
      
      \vspace{-10pt}
      \label{dynamic-memory}
    \end{figure}

\paragraph{\textbf{Memory Interval.}}
    In order to avoid excessive memory redundancy in the same scene, 
    we still choose to trigger the storage of every N frames, but the frame must meet our proposed principles. 
    Otherwise, it will be deferred to the next frame to trigger. 
    As \Cref{memory interval} shows, it is not the smaller the memory interval, the better the segmentation effect. 
    The reason is that although the smaller memory interval means more reference frames, excessive redundancy in the same scene affects the matching process.
    \begin{table}[h]
    \renewcommand\tabcolsep{5pt}
    \footnotesize
    \centering
    \caption{Experiment results of different memory interval on DAVIS2017 val set.}
    \label{memory interval}
    \begin{tabular}{cccc}
    \toprule[1.5pt]
    Memory Interval & $\mathcal{J}$ &$\mathcal{F}$ & $\mathcal{J}\&\mathcal{F}$ \\ \midrule
    3               & 82.3 & 88.1 & 85.2 \\
    5               & \textbf{82.7} & \textbf{88.6} & \textbf{85.7} \\
    7               & 81.6 & 87.8 & 84.7 \\ \bottomrule[1.5pt]
    \end{tabular}
    \end{table}


%
%